\documentclass[manuscript]{aamas} 

\usepackage{latexsym}
\usepackage{enumitem}
\usepackage{flushend}
\usepackage{times}
\usepackage{soul}
\usepackage{url}
\usepackage{hyperref}
\usepackage[utf8]{inputenc}
\usepackage{graphicx}
\usepackage{amsmath}
\usepackage{amsthm}
\usepackage{booktabs}
\usepackage{algorithm}
\usepackage{algorithmic}
\usepackage[switch]{lineno}

\usepackage{makecell}
\usepackage{color, colortbl}
\usepackage{multirow}

\definecolor{grayline}{RGB}{244,246,246}

\usepackage[textsize=tiny
            ]{todonotes}

\usepackage{listings}
\usepackage{xcolor}

\usepackage{balance}

\colorlet{punct}{red!60!black}
\definecolor{background}{HTML}{EEEEEE}
\definecolor{delim}{RGB}{20,105,176}
\colorlet{numb}{magenta!60!black}

\lstdefinelanguage{json}{
    basicstyle=\small\ttfamily,
    numberstyle=\scriptsize,
    stepnumber=1,
    numbersep=5pt,
    showstringspaces=false,
    breaklines=true,
    literate=
     *{0}{{{\color{numb}0}}}{1}
      {1}{{{\color{numb}1}}}{1}
      {2}{{{\color{numb}2}}}{1}
      {3}{{{\color{numb}3}}}{1}
      {4}{{{\color{numb}4}}}{1}
      {5}{{{\color{numb}5}}}{1}
      {6}{{{\color{numb}6}}}{1}
      {7}{{{\color{numb}7}}}{1}
      {8}{{{\color{numb}8}}}{1}
      {9}{{{\color{numb}9}}}{1}
      {:}{{{\color{punct}{:}}}}{1}
      {,}{{{\color{punct}{,}}}}{1}
      {\{}{{{\color{delim}{\{}}}}{1}
      {\}}{{{\color{delim}{\}}}}}{1}
      {[}{{{\color{delim}{[}}}}{1}
      {]}{{{\color{delim}{]}}}}{1},
}

\newcommand{\BibTeX}{B\kern-.05em{\sc i\kern-.025em b}\kern-.08em\TeX}

\newcommand{\system}{{\sc IntentRec4Maps}}
\providecommand\probability[1]{\ensuremath{\mathbb{P}({#1})}}

\acmSubmissionID{34}

\title[Intention Recognition in Real-Time Interactive Navigation Maps]{\LARGE Intention Recognition in Real-Time Interactive Navigation Maps}

\author{Peijie Zhao}
\affiliation{
  \institution{University of Manchester and CUHK}
  \country{United Kingdom and Hong Kong SAR}}

\author{Zunayed Arefin}
\affiliation{
  \institution{University of Manchester}
  \country{United Kingdom}}

\author{Felipe Meneguzzi}
\affiliation{
  \institution{University of Aberdeen and PUCRS}
  \country{United Kingdom and Brazil}}
\email{felipe.meneguzzi@abdn.ac.uk}

\author{Ramon Fraga Pereira}
\affiliation{
  \institution{University of Manchester and UFRGS}
  \country{United Kingdom and Brazil}}
\email{ramon.fragapereira@manchester.ac.uk}

\begin{abstract}

\section*{Abstract}

In this work, we develop \system, a system to recognise users' intentions in interactive maps for real-world navigation. 
\system~uses the Google Maps Platform as the real-world interactive map, and a very effective approach for recognising users' intentions in real-time. 
We experiment and showcase the recognition process of \system~using two different \textit{Path-Planners} and a \textit{Large Language Model} (LLM) as a path-planner via prompts.

\end{abstract}

\begin{document}




\maketitle 


\section{Introduction}

Real-time interactive navigation maps are crucial for us and have become integral tools in our daily lives. Interactive maps use \textit{Global Positioning System} (GPS) technology to provide accurate and real-time location information, being particularly useful for mobile applications, enabling users to track their location and receive turn-by-turn directions for walking, driving, or public transportation.
Existing interactive navigation maps include \textit{markers} for various location points of interest, such as restaurants, gas stations, hotels, etc. This helps users to plan their routes based on their interests or needs during a journey.
A noteworthy functionality of interactive navigation maps, such as Google Maps, Apple Maps, Waze, MapBox, etc, is the use of \textit{Path-Planning} algorithms~\cite{Dijkstra59,AStar_HartNR68,Path_GaoFSKYL14} (along with \textit{heuristics}) to generate optimal routes for users. 
This functionality can be further enriched by integrating real-time traffic data, historical traffic patterns, and various other real-world factors that could be used to ensure the generation of optimal routes.
Nevertheless, as far as we are aware, the existing interactive navigation maps platforms do not have and employ the functionality of \textit{Real-Time Intention Recognition} \cite{AAMAS_PP_MastersS17,JAIR_MastersS19,IJCNN_HuXSTY21}, at least from the user perspective and standpoint.


Empowering centralised systems with the ability to recognise the intended locations of users (either driving, walking, or cycling) could be beneficial to monitor and track resources in a more effective way, such as vehicles or personnel, packages to be delivered, etc, and is of significant importance for logistics, fleet management, or any scenario in which asset movement needs to be closely monitored. 


In this work, we develop \system, a system that is able to recognise users' intended location points in interactive maps for real-world navigation. 
As the interactive navigation map platform, we use the Google Maps Platform, as it provides several useful \textit{Application Programming Interfaces} (API), such as \textit{Maps Embed} API, \textit{Directions} API, \textit{Geocoding} API, etc. For recognising users' intentions, we equip \system~with a real-time recognition approach called \textit{Mirroring}~\cite{ACS_vered2016online}.
\textit{Intention Recognition} has been employed in practice in several distinct applications and scenarios, such as digital games \cite{IJCAI_MinMRLL16}, real-time recognition of culinary recipes in video streams \cite{granada2017hybrid},
planning and decision-making advisor \cite{IBM_IJCAI_Sohrabi0HUF18}, 
agents' behaviour explanation \cite{IJCAI_Jones_ChakrabortiFTDS18}, behavioural cues for recognising human's intentions \cite{AAMAS_SinghMNSVV18}, intention recognition in  latent space images \cite{Demo_ICAPS_Amado2019}, and intent recognition of pedestrians and cyclists via 2D pose estimation~\cite{IEEE_Intent2D_2020}.
To the best of our knowledge, our system \system~stands as a pioneering by employing (real-time) intention recognition in interactive maps for real-world navigation.

We empirically evaluate the efficiency of \system~in complex recognition problems with different numbers of intentions and observations using two different symbolic \textit{Path-Planners} and a \textit{Large Language Model} (LLM). We showcase our system \system~using several examples in a video that can be viewed on \url{https://youtu.be/Nf8g9dxqvFw}.

\section{\system}

\subsection{System Overview}

\begin{figure}[t!]
    \centering
    \includegraphics[scale=0.55]{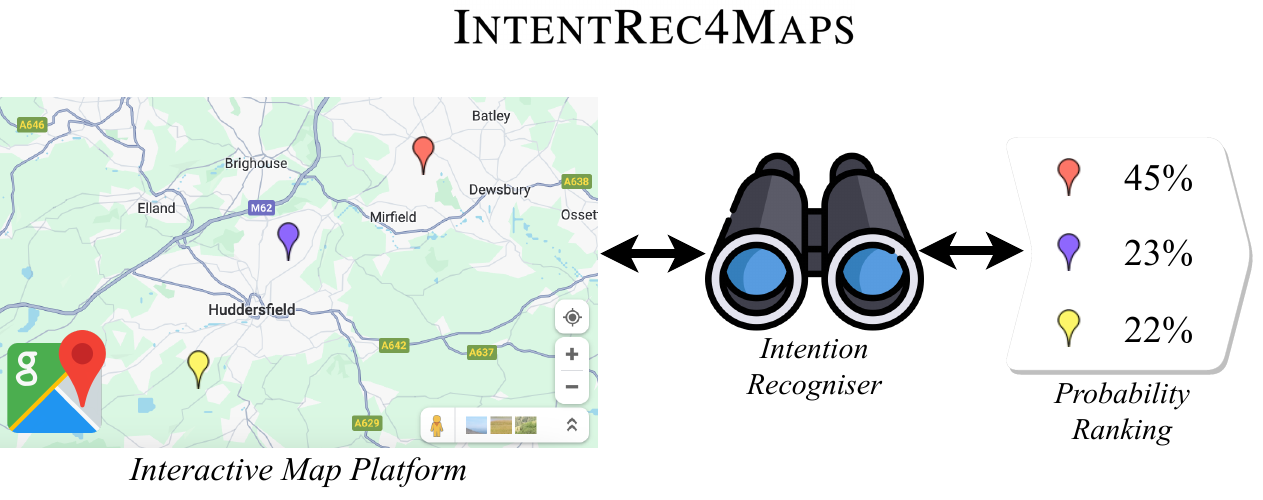}
    \caption{\system~Overview.}
    \label{fig:overview}
\end{figure}

\system~is composed of two main components (Figure~\ref{fig:overview}): the \textit{Interactive Map Platform} component (Section~\ref{sec:interactive-map}), the component that relies on the Google Maps Platform and its APIs to display the interactive map for real-time intention recognition; and the \textit{Real-Time Intention Recognition} component (Section~\ref{sec:recognition}), the component that performs real-time recognition according to a set of input information (possible intentions, observations, etc) and providing a probability distribution reflecting the most and least likely intentions in response to received observations. 
We next explain in detail how these two components work.

\subsection{Interactive Map Platform}\label{sec:interactive-map}

We use the Google Maps Platform as the interactive map, as it provides a very robust set of APIs for real-world navigation.
The environment where \system~performs intention recognition is represented by the road network with \textit{location points of interest} (denoted as $\tt loc$), and the \textit{state-space} consists of the geographical locations on the map. A \textit{location point} is represented using \textit{latitude} and \textit{longitude} coordinates, e.g., $\texttt{loc} = \langle 53.483959,-2.244644 \rangle$.
Actions are represented by moving from one location point to another, following a specific route (or path) according to a transportation mode (e.g., walking, driving, cycling, public transport). We define that a route $\pi = \{ \texttt{loc}_1, ... \texttt{loc}_n \}$ is a sequence of \textit{latitude} and \textit{longitude} coordinates that achieves a specific location point from an initial location point.
Our system \system~relies on the Earth's sphere space (defined in the previous paragraph) to define an \textit{Intention Recognition} problem in real-world navigation maps. We formally define an \textit{Intention Recognition} problem in real-world navigation maps as a tuple $\langle \mathcal{M}, init_{\tt loc}, \mathcal{I}, Obs \rangle$, where: $\mathcal{M}$ is the \textit{real-world map environment}, represented by a road network with location points as latitude and longitude; $init_{\tt loc}$ represents the \textit{initial location point} as latitude and longitude; $\mathcal{I} = \{\texttt{loc}_1, ... \texttt{loc}_n\}$ is a set of possible intended \textit{location points} that an observed user may aim to achieve; and $Obs$ is a sequence of \textit{observations} (represented as latitude and longitude coordinates) that the system observes incrementally, representing a sequence of observed coordinates for achieving an \textit{intended location point} $\texttt{loc}^{*} \in \mathcal{I}$.
An ``ideal solution'' for this problem is recognising (as \textit{top-1} intention in the probability ranking) the \textbf{correct} \textit{intended location point} $\texttt{loc}^{*} \in \mathcal{I}$ (which is \textbf{unknown} from the recognition system's perspective) that an observed user aims to achieve for every observation point $o$ in the observation sequence $Obs$.

\begin{figure}[t!]
    \centering
    \includegraphics[scale=0.65]{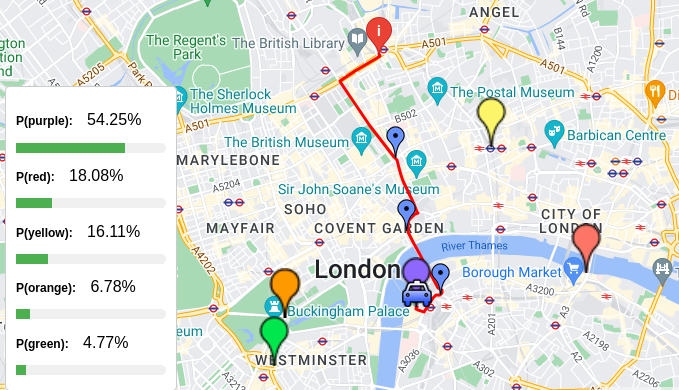}
    \caption{\system~Recognition Process.}
    \label{fig:example-recognition}
\end{figure}

Listing~\ref{lst:recognition-input} shows an example of how we encode an \textit{Intention Recognition} problem using the text encoding of JSON (JavaScript Object Notation). This recognition problem example is depicted visually in Figure~\ref{fig:example-recognition}.

\begin{lstlisting}[language=json,firstnumber=1,caption=\textit{Intention Recognition Problem} example using JSON.,captionpos=b,label={lst:recognition-input}]
  [{
    "problem_id": "1.5.3",
    "init": "Kensington Palace London",
    "intent_location": "Tower Bridge London",
    "intentions": [
      "London Bridge",
      "Univeristy of London",
      "Buckingham Palace London",
      "Tower Bridge London",
      "Farringdon Station London"
    ],
    "observations": [
      [51.502179,-0.1746681],
      [51.511215,-0.0732266],
      [51.509575,-0.0734642]
    ]
  }]
\end{lstlisting}

\subsection{Real-Time Intention Recognition}\label{sec:recognition}

\system~uses a well-known online recognition approach called \textit{Mirroring}~\cite{ACS_vered2016online,AAAI_kaminka2018plan}, a model-based recognition approach~\cite{masters2021_IJCAI_survey,meneguzzi2021_IJCAI_survey}.
\citeauthor{ACS_vered2016online}~\cite{ACS_vered2016online} developed \textit{Mirroring} by arguing that we humans tend to infer other people's intentions by ``\textit{mirroring}'' their \textit{observed} behaviour with some ``\textit{optimal} (\textit{ideal})'' expected behaviour. We adapt the concept of \textit{Mirroring} for real-world navigation maps, and compute two types of routes for the observed user, as follows. We first compute an \textbf{ideal route} $\pi$ from the initial location $init_{\tt loc}$ for every location point $\texttt{loc}$ in the set of possible intentions $\mathcal{I}$. After that, we compute what we call \textbf{observation route} $\pi_{Obs}$, a route that complies with the observations in $Obs$, and is computed from the initial $init_{\tt loc}$ complying with the observations $Obs$ and then achieving the each of the possible intentions $\mathcal{I}$. Thus, for every possible intended location in $\mathcal{I}$, we compare its ideal route $\pi$ with its observation route $\pi_{Obs}$ and compute a \textit{score} (denoted as $\epsilon$).
The score $\epsilon$ is a value between $0$ and $1$, i.e., $ 0 \leq \epsilon \leq 1$, and represents how ``compliant'' (assuming optimal routes~\cite{AAMAS_MastersS19}) the sequence of observations $Obs$ from the agent's behaviour is to a route $\pi$ for achieving a location point of interest, i.e., the closer $\epsilon$ is to zero for location point of interest, the more likely such a location point interest is the intended one. 
We adapt the probabilistic framework of \citeauthor{RamirezG_AAAI2010}~\cite{RamirezG_AAAI2010} to compute a posterior probability distribution for every location point $\texttt{loc}$ in $\mathcal{I}$ using the score $\epsilon$, as formalised as follows:

{
\begin{align*}
\label{eq:posterior}
\probability{loc \mid Obs} = \eta \cdot \probability{loc} \cdot \probability{Obs \mid loc}
\end{align*}
}

\noindent where $\eta$ is normalisation factor, $\probability{loc}$ is a prior probability for a location point, and  $\probability{Obs \mid loc}$ is the probability of the observations $Obs$ for a location point.
We compute $\probability{Obs \mid loc}$ using the score $\epsilon$, namely, $\probability{Obs \mid loc} = [1 + (1-\epsilon)]^{-1}$.
The computation of $\probability{Obs \mid loc}$ involves comparing the routes $\pi$ and $\pi_{Obs}$ to compute the score $\epsilon$. Fundamentally, $\epsilon$ estimates the similarity between the routes $\pi$ and $\pi_{Obs}$, point by point, using the \textit{Haversine Formula} implemented in the $\epsilon(\pi, \pi_{Obs})$ function, ensuring a geographically accurate assessment of spatial separation of the points in $\pi$ and $\pi_{Obs}$ in the Earth's sphere space. A threshold is applied to determine when two location points are considered similar according to their spherical distance, allowing for fine-tuning of the similarity comparison based on specific needs. The \textit{similarity distance comparison} between the location points in $\pi$ and $\pi_{Obs}$, is denoted as $\mathtt{}(l_{\pi},l_{\pi_{Obs}}) = $ 

\begin{figure}[t!]
    \centering
    \includegraphics[scale=0.9]{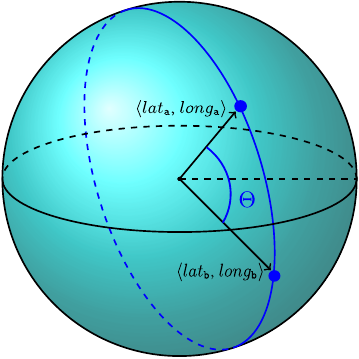}
    \caption{Haversine example.}
    \label{fig:haversine}
\end{figure}

{\scriptsize
\begin{align*}
2 \cdot R \cdot \arcsin\left(\sqrt{\sin^2\left(\frac{\Delta \textit{lat}}{2}\right) + \cos(\textit{lat}_{\pi_{Obs}})\cdot\cos(\textit{lat}_{\pi})\cdot\sin^2\left(\frac{\Delta \textit{long}}{2}\right)}\right)
\end{align*}
}

\noindent where $R$ is the radius of the sphere, $\langle lat_{\pi}, long_{\pi} \rangle$ and $\langle lat_{\pi_{Obs}}, long_{\pi_{Obs}} \rangle$ correspond to the latitude and longitude coordinates for the routes location points in $\pi$ and $\pi_{Obs}$, respectively, and $\Delta lat = |lat_{\pi_{Obs}} - lat_{\pi}|$ and $\Delta long = |long_{\pi_{Obs}} - long_{\pi}|$. Figure~\ref{fig:haversine} exemplifies how we use the \textit{Haversine Formula} to compute the distance between two spatial in the Earth's sphere space.

\section{Experiments and Evaluation}

To experiment and evaluate \system\footnote{The dataset and the code of \system~can be found on the following link: \url{https://github.com/PeijieZ/IntentRec4Maps}.}, we built a real-world navigation recognition dataset with $100$ complex problems, using the Google Maps \textit{Path(Route)-Planner}\footnote{\url{https://developers.google.com/maps/documentation/routes}} to compute the routes and the observations. The recognition problems were built using five different initial locations, and the possible intended location points vary across four groups, ranging from $2$ to $15$ (i.e., $2$, $5$, $10$, and $15$), whereas observation points (extracted from routes using a path-planner) vary across five groups, ranging from $1$ to $15$ (i.e., $1$, $3$, $5$, $10$, and $15$).

Table~\ref{tab:results} shows the results of \system~over the recognition dataset, aggregating the total average overall, grouped by the number of goals, and grouped by the number of observations.
We use three metrics to evaluate \system: \textit{True Positive Rate} (\textit{TPR}), \textit{False Positive Rate} (\textit{FPR}), and \textit{F1-Score} (the trade-off between \textit{TPR} and \textit{FPR}).
We used three experimental setups and tested different ways of extracting routes (paths) for the recognition process: (1) we use the Google Maps \textit{Route-Planner}, as a \textit{Baseline}; (2) we use the MapBox \textit{Route-Planner}\footnote{\url{https://docs.mapbox.com/help/getting-started/directions/}}; and (3) we use a LLM (i.e., ChatGPT4 API\footnote{\url{https://openai.com/blog/gpt-4-api-general-availability}}) as a route-planner, asking directions via prompt.
The \textit{Baseline} represents the \system~and~an observed person using the same navigation system, specifically, the Google Maps API. We use MapBox as an alternative navigation system for the recognition process, making the recognition process more difficult. The rationale for using an LLM as a route-planner is to investigate how ``reliable'' an LLM is when used as a solver for a reasoning/planning process, drawing inspiration from the work of~\cite{LLMsPlanning_2023}.

We can see from the results in Table~\ref{tab:results} that the \textit{Baseline} (grey results) yielded a commendably high recognition accuracy compared to the other recognition setups, which is expected, since it represents the recognition process using the same the navigation system as the user, so it makes the recognition process more efficient. Conversely, Mapbox demonstrated a comparatively lower accuracy than the \textit{Baseline}, which is expected, considering the divergence between the possible extracted routes from Google Maps. Note that Mapbox yielded commendable results when compared with LLM (ChatGPT4).
It is noteworthy that the results for the recognition process using an LLM (ChatGPT4) are the worst (in terms of accuracy) when compared with the other recognition setups. Such results are somewhat expected since LLMs are not explicitly designed for extracting the routes.
\begin{table}[t!]
\centering
\setlength\tabcolsep{3pt}
\renewcommand{\arraystretch}{1}
\Large\selectfont

\begin{tabular}{rrrrrrrrrrrr}
        \toprule
        \multicolumn{1}{l}{} & \multicolumn{3}{c}{Google Maps}                             & \multicolumn{1}{l}{} & \multicolumn{3}{c}{MapBox}                                & \multicolumn{1}{l}{} & \multicolumn{3}{c}{LLM}                                   \\

        \multicolumn{1}{l}{} & \multicolumn{3}{c}{(\textit{Baseline})}                             & \multicolumn{1}{l}{} & \multicolumn{3}{c}{}                                & \multicolumn{1}{l}{} & \multicolumn{3}{c}{}

        \\

        \cmidrule[\heavyrulewidth]{2-4}
        \cmidrule[\heavyrulewidth]{6-8}
        \cmidrule[\heavyrulewidth]{10-12}

        \multicolumn{1}{l}{$|Obs|$} & \multicolumn{1}{c}{\it TPR} & \multicolumn{1}{c}{\it FPR} & \multicolumn{1}{c}{\it F1} & \multicolumn{1}{l}{} & \multicolumn{1}{c}{\it TPR} & \multicolumn{1}{c}{\it FPR} & \multicolumn{1}{c}{\it F1} & \multicolumn{1}{l}{} & \multicolumn{1}{c}{\it TPR} & \multicolumn{1}{c}{\it FPR} & \multicolumn{1}{c}{\it F1}

        \\

        \cmidrule[\heavyrulewidth]{2-4}
        \cmidrule[\heavyrulewidth]{6-8}
        \cmidrule[\heavyrulewidth]{10-12}

        \rowcolor{grayline}1 & \textcolor{gray}{0.90} & \textcolor{gray}{0.15} & \textcolor{gray}{0.61} &  & \textbf{0.20} & \textbf{0.96} & \textbf{0.01} &  & \textbf{0.20} & \textbf{0.96} & \textbf{0.01} \\

        3 & \textcolor{gray}{1.00} & \textcolor{gray}{0.00} & \textcolor{gray}{1.00} &  & \textbf{0.40} & \textbf{0.70} & \textbf{0.15} &  & 0.25 & 0.82 & 0.02 \\

        \rowcolor{grayline}5 & \textcolor{gray}{1.00} & \textcolor{gray}{0.00} & \textcolor{gray}{1.00} &  & \textbf{0.70} & \textbf{0.45} & \textbf{0.20} &  & 0.30 & 0.76 & 0.03 \\

        10 & \textcolor{gray}{1.00} & \textcolor{gray}{0.00} & \textcolor{gray}{1.00} &  & \textbf{0.78} & \textbf{0.40} & \textbf{0.29} &  & 0.35 & 0.78 & 0.04 \\

        \rowcolor{grayline}15 & \textcolor{gray}{1.00} & \textcolor{gray}{0.00} & \textcolor{gray}{1.00} &  & \textbf{0.78} & \textbf{0.40} & \textbf{0.29} &  & 0.40 & 0.78 & 0.06 \\
        \bottomrule
\end{tabular}
\caption{Experiments results of \system.}
\label{tab:results}
\end{table}

\section{Conclusions}

We developed \system, a novel recognition system that is able to recognise users' intentions in interactive maps for real-world navigation. Our system employees the \textit{Haversine Formula} to compute distances between locations in the Earth's sphere space.
As future work, we aim to extend \system~and implement other recognition features and functionalities, such as dealing with irrational~\cite{AAMAS_MastersS19} and possibly adversarial behaviour, and noisy and spurious observations. Furthermore, we aim to assess the recognition accuracy of \system~when dealing with long-distance intended location points.


\begin{acks}
  We would like to thank the referees for their valuable comments. 
  This study was part funded by the Coordenação de Aperfeiçoamento de Pessoal de Nível Superior -- Brasil (CAPES) -- Finance Code 001.
\end{acks}


\bibliographystyle{ACM-Reference-Format} 
\bibliography{references}


\end{document}